\documentclass[11pt,a4paper]{article}
\usepackage{graphicx} 

\usepackage{amsmath,amsfonts,amssymb}
\usepackage{cite,url,algorithm}
\usepackage{color,subfigure}

\graphicspath{{figures/}}

\usepackage{enumitem}

\newcommand{\matrice}[2]{\left[\hspace*{-.1cm}\ba{#1} #2 \ea\hspace*{-.1cm}\right]}
\newcommand{\ba}[1]{\begin{array}{#1}}
    \newcommand{\ea}{\end{array}}
\newcommand{\beqarno}{\begin{eqnarray*}}
    \newcommand{\eeqarno}{\end{eqnarray*}}
\newcommand{\beqar}{\begin{eqnarray}}
    \newcommand{\eeqar}{\end{eqnarray}}
\newcommand{\st}{\mathop{\rm s.t.}\nolimits}
\newcommand{\rr}{{\mathbb R}}
\newcommand{\FF}{\mathcal{F}}
\newcommand{\XX}{\mathcal{X}}
\newcommand{\diag}{\mathop{\rm diag}\nolimits}
\newcommand{\QED}{\hfill\mbox{\rule[0pt]{1.5ex}{1.5ex}}\vspace{0em}\par\noindent}
\newcommand{\smallmat}[1]{\left[ \begin{smallmatrix}#1 \end{smallmatrix} \right]}
\ifx\thelemma\undefined
\newtheorem{lemma}{Lemma}
\fi
\ifx\proof\undefined
\newcommand{\proof}{\noindent{\em Proof}.~}
\fi

\begin{document}
    
    \title{Global optimization via inverse distance weighting\\ and radial basis functions}
   
    \author{Alberto Bemporad\thanks{A. Bemporad is with the IMT School for Advanced Studies Lucca, Italy.
            Email: \texttt{alberto.bemporad@imtlucca.it}.}\\\vspace*{1cm}
    }
  
    \maketitle

\begin{abstract}
Global optimization problems whose objective function is 
expensive to evaluate can be solved effectively by recursively fitting 
a \emph{surrogate function} to function samples and 
minimizing an \emph{acquisition function} to generate
new samples. The acquisition step trades off between seeking for a new
optimization vector where the surrogate is minimum (\emph{exploitation} of the surrogate) and
looking for regions of the feasible space that have not yet been visited and that may 
potentially contain better values of the objective function
(\emph{exploration} of the feasible space). This paper proposes a new global optimization algorithm that uses a combination of inverse distance weighting (IDW)
and radial basis functions (RBF) to construct the acquisition function.
Rather arbitrary constraints that are simple to evaluate can be easily taken into account. 
Compared to Bayesian optimization,
the proposed algorithm, that we call GLIS (GLobal minimum using Inverse distance weighting and Surrogate radial basis functions), is competitive and computationally lighter, 
as we show in a set of benchmark global optimization and hyperparameter tuning problems.
MATLAB and Python implementations of GLIS are available at \url{http://cse.lab.imtlucca.it/~bemporad/glis}.
\end{abstract}


\section{Introduction}
Many problems in machine learning and statistics, engineering design, physics, medicine, management science, and in many other fields, require finding a global minimum of
a function without derivative information; see, e.g., the excellent survey on derivative-free optimization~\cite{RS13}. Some of the most successful approaches for derivative-free global optimization include deterministic methods based on recursively splitting the feasible
space in rectangles, such as the DIRECT (DIvide a hyper-RECTangle)~\cite{Jon09}
and Multilevel Coordinate Search (MCS)~\cite{HN99} algorithms, and stochastic methods
such as Particle Swarm Optimization (PSO)~\cite{EK95}, 
genetic algorithms~\cite{Whi94}, and evolutionary algorithms~\cite{HO01}.

The aforementioned methods can be very successful in reaching a global minimum without any assumption on convexity and smoothness of the function, but may result in evaluating the function a large
number of times during the execution of the algorithm. In many problems, however, the objective function is a black box that can be very time-consuming to evaluate. For example, in hyperparameter tuning of machine learning algorithms, one needs to run a large set of training tests per hyperparameter choice; in structural engineering design, testing the resulting mechanical property corresponding to a given choice of parameters may involve several hours for computing solutions to partial differential equations; in control systems design, testing a combination of controller parameters involves running a real closed-loop experiment, which is time consuming and costly.
For this reason, many researchers have been studying algorithms for black-box global optimization
that aim at minimizing the number of function evaluations by replacing the function to minimize with a \emph{surrogate} function~\cite{Jon01}. The latter is obtained by sampling the
objective function and interpolating the samples with a map that, compared to the original function, is very cheap to evaluate. The surrogate is then used to solve a (much cheaper) global optimization problem that decides the new point the original function must be evaluated. A better-quality surrogate is then created by also exploiting the new sample and the procedure is iterated. For example, quadratic surrogate functions are used in the well known global optimization method NEWUOA~\cite{Pow06}.

As underlined by several authors (see, e.g.,~\cite{Jon01}), purely minimizing the surrogate function may lead to converge to a point that is not the global minimum of the black-box function. 
To take into account the fact that the surrogate and the true objective function differ from each other in an unknown way,
the surrogate is typically augmented by an extra term that takes into
account such an uncertainty. The resulting \emph{acquisition function}
is therefore minimized instead for generating a new sample of the optimization vector, trading off
between seeking for a new
vector where the surrogate is small and
looking for regions of the feasible space that have not yet been visited.

Bayesian Optimization (BO) is a popular class of global optimization methods based on surrogates that, by modeling the black box function as a Gaussian process, enables one to quantify in statistical terms the discrepancy between the two functions, an information that is taken into account to drive the search. BO has been studied since the sixties in global optimization~\cite{Kus64} and in geostatistics~\cite{Mat63} under the name of Kriging methods; it become popular to solve
problems of Design and Analysis of Computer Experiments (DACE)~\cite{SWMW89}, see for instance
the popular Efficient Global Optimization (EGO) algorithm~\cite{JSW98}. It is nowadays very popular in machine learning for tuning hyperparameters of different  algorithms~\cite{BCD10,SLA12,SSWAD15,FKESBH19}. 

Motivated by learning control systems from data~\cite{PFFB19} and self-calibration of optimal control parameters~\cite{FPB20}, in this paper
we propose an alternative approach to solve global optimization problems in which the objective function is expensive to evaluate that is based on Inverse Distance Weighting (IDW) interpolation~\cite{She68,JK11}
and Radial Basis Functions (RBFs)~\cite{Har71,MGTM07}.  
The use of RBFs for solving global optimization problems was
already adopted in~\cite{Gut01,CN18}, in which the acquisition function is constructed by introducing a ``measure of bumpiness''. The author of~\cite{Gut01} shows that such a measure has a relation with the probability of hitting a lower value than a given threshold of the underlying function, as used in Bayesian optimization. RBFs were also adopted in~\cite{RS05}, with additional
constraints imposed to make sure that the feasible set is adequately explored. In this paper we use a different acquisition function based on two components: an estimate of the confidence interval associated with RBF interpolation as suggested
in~\cite{JK11}, and a new measure based on inverse distance weighting that is totally independent of the underlying black-box function
and its surrogate. Both terms aim at exploring the domain of the optimization vector.
Moreover, arbitrary constraints that are simple to evaluate are also taken into account, as they can be easily imposed during the minimization of the acquisition function. 

Compared to Bayesian optimization, our non-probabilistic approach to global optimization 
is very competitive, as we show in a set of benchmark global optimization problems
and on hyperparameter selection problems, and also computationally lighter
than off-the-shelf implementations of BO.

A preliminary version of this manuscript was made available in~\cite{Bem19} and later
extended in~\cite{BP19} to solve preference-based optimization problems. MATLAB and a Python implementations of the proposed approach and of the one of~\cite{BP19} are available for download at \url{http://cse.lab.imtlucca.it/~bemporad/glis}. For an application
of the GLIS algorithm proposed in this paper to learning optimal
calibration parameters in embedded model predictive control applications
the reader is referred to~\cite{FPB20}.

The paper is organized as follows. After stating the global optimization problem we want to solve in Section~\ref{sec:prob_formulation}, Sections~\ref{sec:surrogate} and~\ref{sec:acquisition} deal with the construction of the surrogate and acquisition functions, respectively.
The proposed global optimization algorithm is detailed in Section~\ref{sec:algorithm}
and several results are reported in Section~\ref{sec:results}. Finally, some conclusions
are drawn in Section~\ref{sec:conclusions}.

\section{Problem formulation}
\label{sec:prob_formulation}
Consider the following constrained global optimization problem
\begin{equation}
    \ba{rl}
    \min_x& f(x)\\
    \st & \ell\leq x\leq u\\
        & x\in\XX 
    \ea
\label{eq:glob-opt}
\end{equation}
where $f:\rr^n\to\rr$ is an arbitrary function of the optimization vector $x\in\rr^n$, $\ell,u\in\rr^n$ are vectors of lower and upper bounds, and $\XX\subseteq\rr^n$ imposes further arbitrary constraints on $x$. Typically $\XX=\{x\in\rr^n:\ g(x)\leq 0\}$, where the vector function $g:\rr^n\to\rr^q$ defines inequality constraints, 
with $q=0$ meaning that no inequality constraint is enforced; 
for example, linear inequality constraints are defined by setting $g(x)=Ax-b$, with $A\in\rr^{q\times n}$, $b\in\rr^q$, $q\geq 0$. 
We are particularly interested in problems as in~\eqref{eq:glob-opt} such that $f(x)$ is expensive to evaluate
and its gradient is not available, while the condition $x\in\XX$ 
is easy to evaluate. Although not comprehensively addressed in this paper, we will show that our approach
also tolerates noisy evaluations of $f$, that is if we measure $y=f(x)+\varepsilon$ instead of $f(x)$, 
where $\varepsilon$ is an unknown quantity. We will not make any assumption on $f$, $g$, and $\varepsilon$.
In~\eqref{eq:glob-opt} we do not include possible linear equality
constraints $A_{e}x=b_e$, as they can be first eliminated by reducing 
the number of optimization variables.

\section{Surrogate function}
\label{sec:surrogate}
Assume that we have collected a vector $F=[f_1\ \ldots\ f_N]'$
of $N$ samples $f_i=f(x_i)$ of $f$, $F\in\rr^N$ at corresponding points $X=[x_1\ \ldots\ x_N]'$, $X\in\rr^{N\times n}$, with $x_i\neq x_j$, $\forall i\neq j$,
$i,j=1,\ldots,N$. We consider next two types of surrogate functions, namely 
\emph{Inverse Distance Weighting} (IDW) functions~\cite{She68,JK11} and
\emph{Radial Basis Functions} (RBFs)~\cite{Gut01,MGTM07}.

\subsection{Inverse distance weighting functions}
Given a generic new point $x\in\rr^n$ consider the vector function of squared Euclidean distances $d^2:\rr^{n\times n}\to\rr^N$
\begin{equation}
    d^2(x,x_i)=(x_i-x)'(x_i-x),\ i=1,\ldots,N
\label{eq:distance}
\end{equation}
In standard IDW functions~\cite{She68} the weight functions 
$w_i:\rr^n\setminus\{x_i\}\to\rr$ are defined by the inverse squared distances
\begin{subequations}
\begin{equation}
    w_i(x)=\frac{1}{d^2(x,x_i)}
\label{eq:w-IDW-basic}
\end{equation}
The alternative weighting function
\begin{equation}
    w_i(x)=\frac{e^{-d^2(x,x_i)}}{d^2(x,x_i)}
\label{eq:w-IDW}
\end{equation}
\end{subequations}
suggested in~\cite{JK11}  has the advantage of
being similar to the inverse squared distance in~\eqref{eq:w-IDW-basic}
for small values of $d^2$, but makes the effect of points $x_i$ located
far from $x$ fade out quickly due to the exponential term.

By defining for $i=1,\ldots,N$ the following functions $v_i:\rr^n\to\rr$
as
\begin{equation}
    v_i(x)=\left\{\ba{ll}
    1&\mbox{if}\ x=x_i\\
    0&\mbox{if}\ x=x_j,\ j\neq i\\
    \displaystyle{\frac{w_i(x)}{\sum_{j=1}^Nw_j(x)}} & \mbox{otherwise}\ea\right.
\label{eq:v}
\end{equation}
the surrogate function $\hat f:\rr^n\to\rr$
\begin{equation}
\hat f(x)=\sum_{i=1}^Nv_i(x)f_i
\label{eq:fhat-idw}
\end{equation}
is an IDW interpolation of $(X,F)$.

\begin{lemma}
The IDW interpolation function $\hat f$ defined in~\eqref{eq:fhat-idw} enjoys the following properties:
\begin{enumerate}
\item [P1.] $\hat f(x_j)=f_j$, $\forall j=1,\ldots,N$;
\item [P2.] $\min_j\{f_j\}\leq \hat f(x)\leq \max_j\{f_j\}$, $\forall x\in\rr^n$;
\item [P3.] $\hat f$ is differentiable everywhere on $\rr^n$ and in particular $\nabla f(x_j)=0$ for all $j=1,\ldots,N$.
\end{enumerate}
\label{lemma:fhat}
\end{lemma}
The proof of Lemma~\ref{lemma:fhat} is very simple and is reported in Appendix~\ref{app:proofs}.

Note that in~\cite{JK11} the authors suggest to improve the surrogate function by adding
a regression model in~\eqref{eq:fhat-idw} to take global trends into account. In our numerical experiments
we found, however, that adding such a term does not lead to significant improvements of the proposed
global optimization algorithm.

\begin{figure}[t]
\begin{center}{\includegraphics[width=.9\hsize]{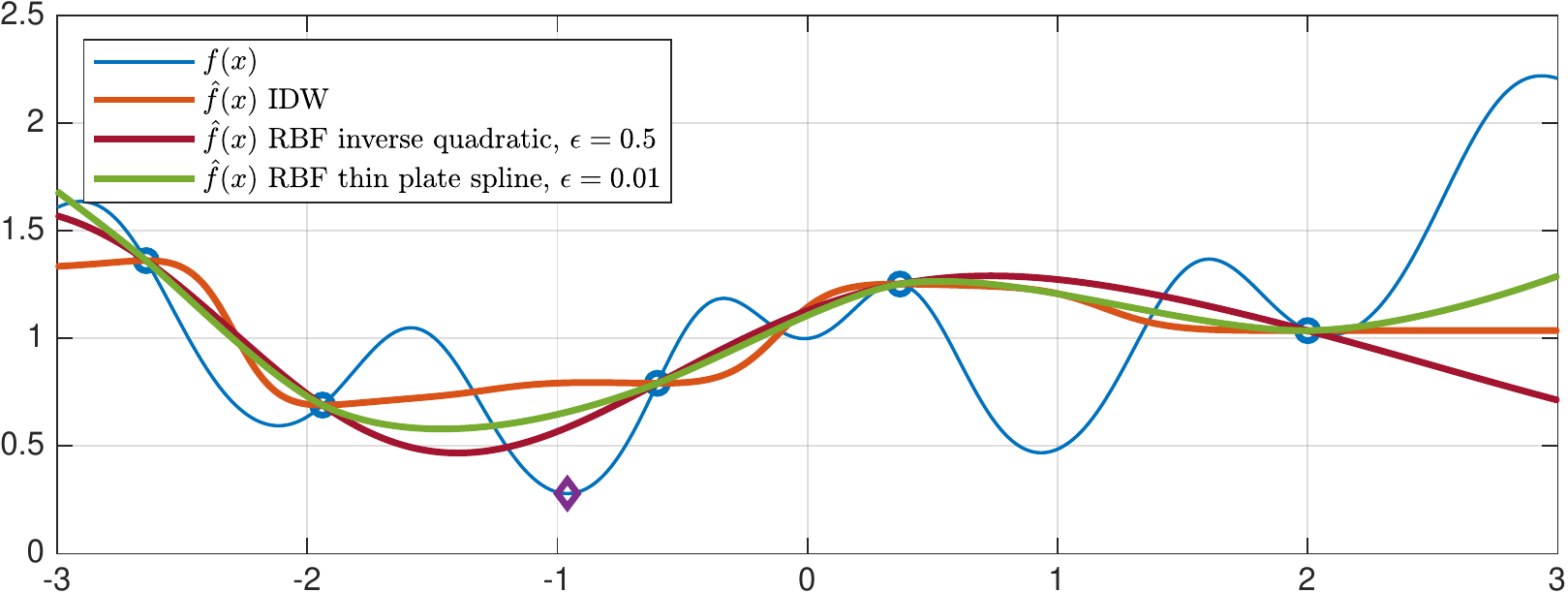}}\end{center}
\caption{A scalar example of $f(x)$ as in~\eqref{eq:f-1d-example} (blue)
sampled at $N=5$ points (blue circles), IDW surrogate $\hat f(x)$ (orange)
with $w_i(x)$ as in~\eqref{eq:w-IDW}, 
RBF inverse quadratic with $\epsilon=0.5$ (red),
RBF thin plate spline surrogate with $\epsilon=0.01$ (green),
global minimum (purple diamond)}
\label{fig:rbf_vs_idw}
\end{figure}

A one-dimensional example of the IDW surrogate $\hat f$ sampled at five different points of 
the scalar function
\begin{equation}
    f(x)=\left(1+\frac{x\sin(2x)\cos(3x)}{1+x^2}\right)^2+\frac{x^2}{12}+\frac{x}{10}
\label{eq:f-1d-example}
\end{equation}
is depicted in Figure~\ref{fig:rbf_vs_idw}. The global optimizer is $x^*\approx-0.9599$
corresponding to the global minimum $f(x^*)\approx0.2795$.

\subsection{Radial basis functions}
A possible drawback of the IDW function $\hat f$ defined in~\eqref{eq:fhat-idw}
is due to property P3: As the number $N$ of samples increases, the surrogate
function tends to ripple, having its derivative to always assume zero value
at samples. An alternative is to use a radial basis function (RBF)~\cite{Gut01,MGTM07}
as a surrogate function. These are defined by setting
\begin{equation}
    \hat f(x)=\sum_{i=1}^N\beta_i\phi(\epsilon d(x,x_i))
\label{eq:rbf}
\end{equation}
where $d:\rr^{n\times n}\to\rr$ is the function defining the Euclidean distance
as in~\eqref{eq:distance}, $d(x,x_i)=\|x-x_i\|_2$, 
$\epsilon>0$ is a scalar parameter, $\beta_i$ are coefficients to be determined as explained below, and $\phi:\rr\to\rr$ is a RBF. Popular examples of RBFs are
\begin{equation}
    \ba{llll}
    \phi(\epsilon d)=\frac{1}{1+(\epsilon d)^2}& \mbox{\small (inverse quadratic)} &
    \phi(\epsilon d)=e^{-(\epsilon d)^2}& \mbox{\small (Gaussian)}\\[1em]
    \phi(\epsilon d)=\sqrt{1+(\epsilon d)^2}& \mbox{\small (multiquadric)} & 
    \phi(\epsilon d)=(\epsilon d)^2\log(\epsilon d)& \mbox{\small (thin plate spline)}\\[1em]
    \phi(\epsilon d)=\epsilon d& \mbox{\small (linear)} &
    \phi(\epsilon d)=\frac{1}{\sqrt{1+(\epsilon d)^2}}& \mbox{\small (inverse multiquadric)}
    \ea 
\label{eq:RBF-examples}
\end{equation}
The coefficient vector $\beta=[\beta_1\ \ldots\ \beta_N]'$  is obtained by 
imposing the interpolation condition
\begin{equation}
    \hat f(x_i)=f_i,\ i=1,\ldots,N
\label{eq:RBF-interp}
\end{equation}
Condition~\eqref{eq:RBF-interp} leads to solving the linear system
\begin{subequations}
\begin{equation}
    M\beta = F
\label{eq:RBF-system}
\end{equation}
where $M$ is the $N\times N$ symmetric matrix whose $(i,j)$-entry is
\begin{equation}
    M_{ij}=\phi(\epsilon d(x_i,x_j))
\label{eq:RBF-matrix}
\end{equation}
with $M_{ii}=1$ for all the RBF type listed in~\eqref{eq:RBF-examples}
but the linear and thin plate spline, for which $M_{ii}=\lim_{d\rightarrow 0}\phi(\epsilon d)=0$.
Note that if function $f$ is evaluated at a new sample $x_{N+1}$, matrix $M$ only requires adding the last row/column obtained by computing $\phi(\epsilon d(x_{N+1},x_j))$
for all $j=1,\ldots,N+1$.

As highlighted in~\cite{Gut01,Jon01}, matrix $M$ might be singular, even if $x_i\neq x_j$ for all $i\neq j$. To prevent issues due to a singular $M$,~\cite{Gut01,Jon01} suggest
using a surrogate function given by the sum of a RBF and a polynomial function of a certain degree. To also take into account unavoidable numerical issues when
distances between sampled points get close to zero, which will easily happen as new
samples are added towards finding a global minimum, in this paper we suggest instead to use a singular value decomposition (SVD) $M=U\Sigma V'$ of $M$ \footnote{
Matrices $U$ and $V$ have the same columns, modulo a change a sign. Indeed, as $M$ is symmetric, we could instead solve the symmetric eigenvalue problem $M=T'\Lambda T$, $T'T=I$, which gives $\Sigma_{ii}=|\Lambda_{ii}|$, and set $U=V=T'$. As $N$ will be typically be small, we neglect computational advantages and adopt here SVD decomposition.}. By neglecting singular values below a certain positive threshold $\epsilon_{\rm SVD}$, we can approximate
$\Sigma=\smallmat{\Sigma_1 & 0\\0 & 0}$, where $\Sigma_1$ collects all singular values
$\sigma_i\geq \epsilon_{\rm SVD}$, and accordingly split $V=[V_1\ V_2]$, $U=[U_1\ U_2]$
so that
\begin{equation}
    \beta=V_1\Sigma_1^{-1}U_1'F
\label{eq:betaRBS}
\end{equation}
\label{eq:rbf_solve}
\end{subequations}

\begin{figure}[t]
\begin{center}{\includegraphics[width=.9\hsize]{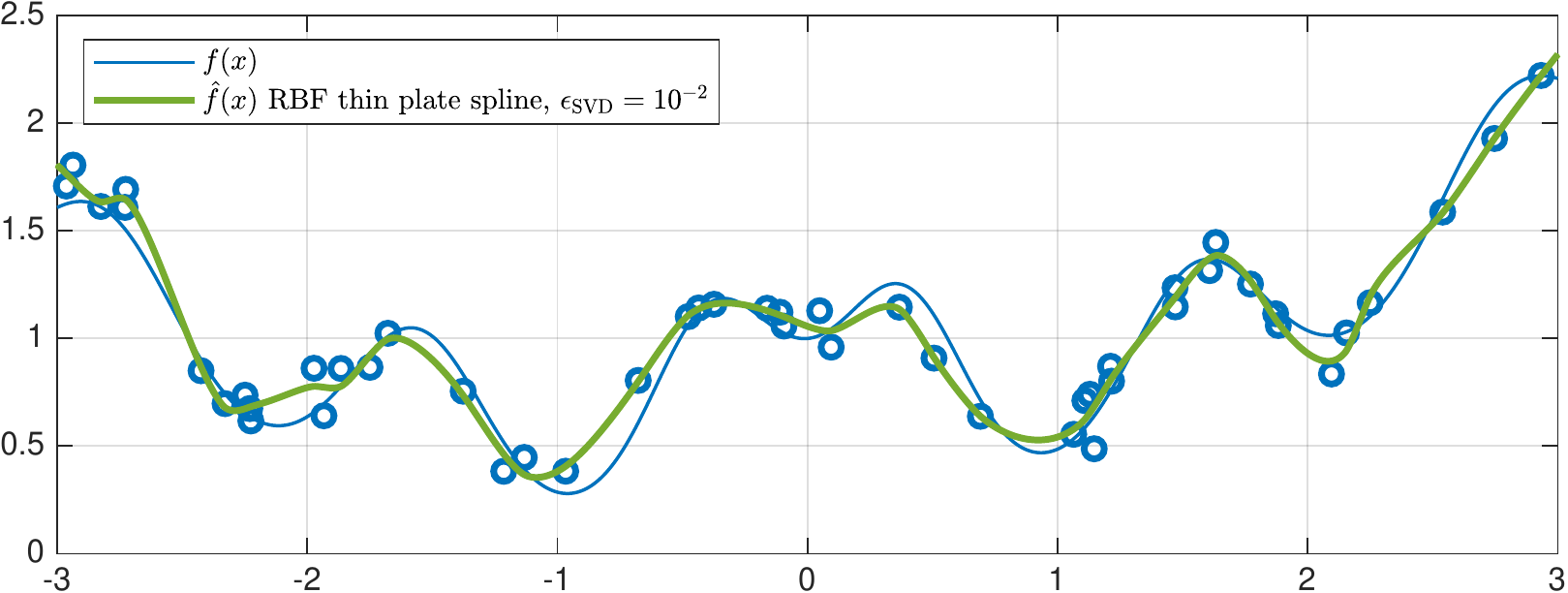}}\end{center}
\caption{Function $f(x)$ as in~\eqref{eq:f-1d-example} is sampled 50 times,
with each sample corrupted by noise $\varepsilon\sim{\mathcal N}(0,10^{-2})$ (blue).
The RBF thin plate spline surrogate with $\epsilon=0.01$ (green)
is obtained by setting $\epsilon_{\rm SVD}=10^{-2}$}
\label{fig:f_rbf_noise}
\end{figure}

The threshold $\epsilon_{\rm SVD}$ turns out to be useful when dealing
with noisy measurements $y=f(x)+\varepsilon$ of $f$. Figure~\ref{fig:f_rbf_noise}
shows the approximation $\hat f$ obtained from 50 samples with $\varepsilon$ normally distributed around zero with standard deviation $0.1$, when $\epsilon_{\rm SVD}=10^{-2}$.

A drawback of RBFs, compared to IDW functions, is that property P2
is no longer satisfied, with the consequence that the surrogate may extrapolate
large values $\hat f(x)$ where $f(x)$ is actually small, and vice versa.
See the examples plotted in Figure~\ref{fig:rbf_vs_idw}.
On the other hand, while differentiable everywhere, RBFs do not necessarily have
zero gradients at sample points as in P3, which is favorable to better approximate
the underlying function with limited samples. For the above reasons, we will mostly
focus on RBF surrogates in our numerical experiments.

\subsection{Scaling}
To take into account that different components $x^j$ of $x$ may have different ranges
$u^j-\ell^j$, we simply rescale the variables in optimization problem~\eqref{eq:glob-opt}
so that they all range in $[-1,1]$. To this end, 
we first possibly tighten the given box constraints $B_{\ell,u}=\{x\in\rr^n: \ell\leq x\leq u\}$ by computing the bounding box $B_{\ell_s,u_s}$ of the set $\{x\in\rr^n:\ x\in\XX 
\}$ and replacing $B_{\ell,u}\leftarrow B_{\ell,u}\cap B_{\ell_s,u_s}$. 
The bounding box $B_{\ell_s,u_g}$ is obtained by solving the following $2n$
optimization problems
\begin{subequations}
\begin{equation}
    \ba{rcl}
    \ell_g^i&=&\min_{x\in\XX
                } e_i'x\\
    u_g^i&=&\max_{x\in\XX
                } e_i'x
    \ea 
\label{eq:BBox}
\end{equation}
where $e_i$ is the $i$th column of the identity matrix, $i=1,\ldots,n$.
In case of linear inequality constraints $\XX=\{x\in\rr^n\ Ax\leq b\}$
, the problems in~\eqref{eq:BBox}
can be solved by linear programming (LP), see~\eqref{eq:ChebyRadius} below.
Since now on, we assume that $\ell,u$ are replaced by
\begin{equation}
    \ba{rcl}
    \ell&\leftarrow&\max\{\ell,\ell_g\}\\
    u&\leftarrow&\min\{u,u_g\}
    \ea
\label{eq:BBoxAb}
\end{equation}
\label{eq:BBox-tighten}%
\end{subequations}
where ``$\min$ ''and ``$\max$'' in~\eqref{eq:BBoxAb} operate component-wise.
Next, we introduce scaled variables $\bar x\in\rr^n$ whose relation with $x$ is
\begin{subequations}
\begin{equation}
    x^j(\bar x)=\bar x^j(u^j-\ell^j)+\frac{u^j+\ell^j}{2}
\label{eq:xy}
\end{equation}
for all $j=1,\ldots,n$ and finally formulate the following scaled 
global optimization problem
\begin{equation}
    \ba{rl}
    \min & f_s(\bar x) \\
    \st & -1\leq \bar x^j\leq 1,\ j=1,\ldots,n\\
        & x\in\XX_s 
    \ea
\label{eq:glob-opt-scaled}
\end{equation}
where $f_s:\rr^n\to\rr$, $\XX_s$ 
are defined as
\[
    \ba{rcl}
    f_s(\bar x)&=&f(x(\bar x))    \\
    \XX_s&=&\{\bar x\in\rr^n:\ x(\bar x)\in\XX\} 
    \ea 
\]
In case $\XX$ is a polyhedron
we have
\begin{equation}
    \XX_s=\{\bar x:\ \bar A\bar x\leq \bar b\}
    \label{eq:gs_Ab_bar}
\end{equation}
where $\bar A$, $\bar b$ are a rescaled version of $A,b$ defined as
\begin{equation}
    \ba{rcl}
       \bar A&=&A\diag(u-\ell)\\
        \bar b&=&b-A(\frac{u+\ell}{2})
    \ea 
\label{eq:Ab_bar}
\end{equation}
\label{eq:scaling}%
\end{subequations}
and $\diag(u-\ell)$ is the diagonal matrix whose diagonal elements
are the components of $u-\ell$. 

Note that, when approximating $f_s$ with $\hat f_s$,
we use the squared Euclidean distances 
\[
    d^2(\bar x,\bar x_i)=\sum_{h=1}^n (\bar x-\bar x_i)^2=\sum_{h=1}^n \left(\theta_h(x^h-x^h_i)\right)^{p_h}
\]
where the scaling factors $\theta^h=u^h-\ell^h$ and $p_h\equiv 2$ are constant.
Therefore, finding a surrogate $\hat f_s$ of $f_s$ in $[-1,1]$ is equivalent
to finding a surrogate $\hat f$ of $f$ under scaled distances.
This is a much simpler scaling approach than computing the scaling factors $\theta^h$
and power $p$ as it is common in stochastic process model approaches
such as Kriging methods~\cite{SWMW89,JSW98}. As highlighted in~\cite{Jon01},
Kriging methods use radial basis functions $\phi(x_i,x_j)=e^{-\sum_{h=1}^n\theta_h|x_i^h-x_j^h|^{p_h}}$,
a generalization of Gaussian RBF functions in which the scaling factors
and powers that are recomputed as the data set $X$ changes.

Note also that the approach adopted in~\cite{BP19} for scaling automatically the surrogate function via cross-validation
could be also used here, as well as other approaches specific for RBFs such as Rippa's method~\cite{Rip99}.
In our numerical experiments we have found that adjusting the RBF parameter $\epsilon$ via cross-validation, while increasing the computational effort, does not provide significant benefit. What
is in fact most critical is the tradeoff between exploitation of the surrogate and exploration of the feasible set, that we discuss in the next section.

\section{Acquisition function}
\label{sec:acquisition}
As mentioned earlier, minimizing the surrogate function to get a new sample
$x_{N+1}$ $=$ $\arg\min\hat f(x)$ subject to $\ell\leq x\leq u$ and $x\in\XX$
, evaluating $f(x_{N+1})$, and iterating over $N$ may easily miss the global minimum of $f$. This is particularly evident when $\hat f$ is the IDW surrogate~\eqref{eq:fhat-idw}, that by Property P2 of Lemma~\ref{lemma:fhat} 
has a global minimum at one of the existing samples $x_i$. Besides \emph{exploiting} the surrogate function $\hat f$, when looking for a new candidate optimizer $x_{N+1}$ it is therefore necessary to add to $\hat f$  a term for \emph{exploring}  areas of the feasible space that have not yet been probed.

\begin{figure}[t]
\begin{center}
\includegraphics[width=.9\hsize]{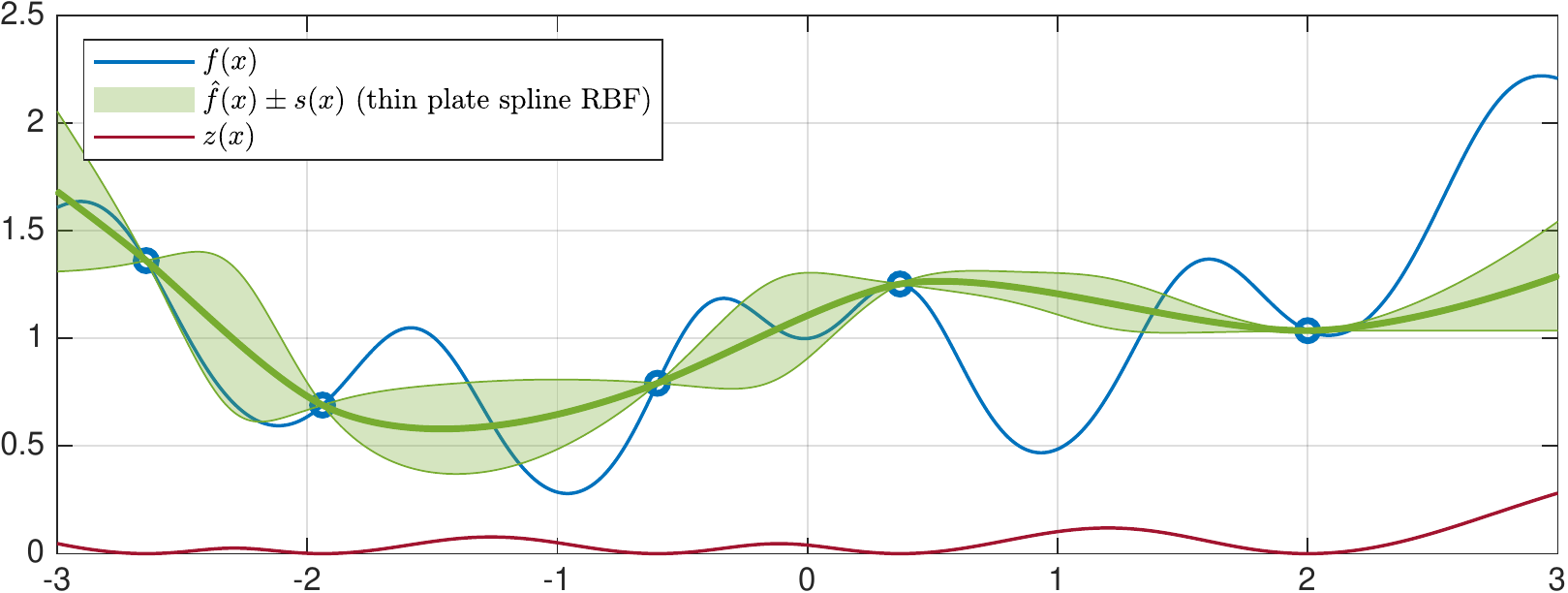}
\end{center}
\caption{Plot of $\hat f(x)\pm s(x)$ and $z(x)$ for the scalar example
as in Figure~\ref{fig:rbf_vs_idw}, with $w_i(x)$ as in~\eqref{eq:w-IDW-basic} and $f$ as in~\eqref{eq:f-1d-example}}
\label{fig:f_s_d}
\end{figure}

\begin{figure}[t]
\begin{center}
\includegraphics[width=.9\hsize]{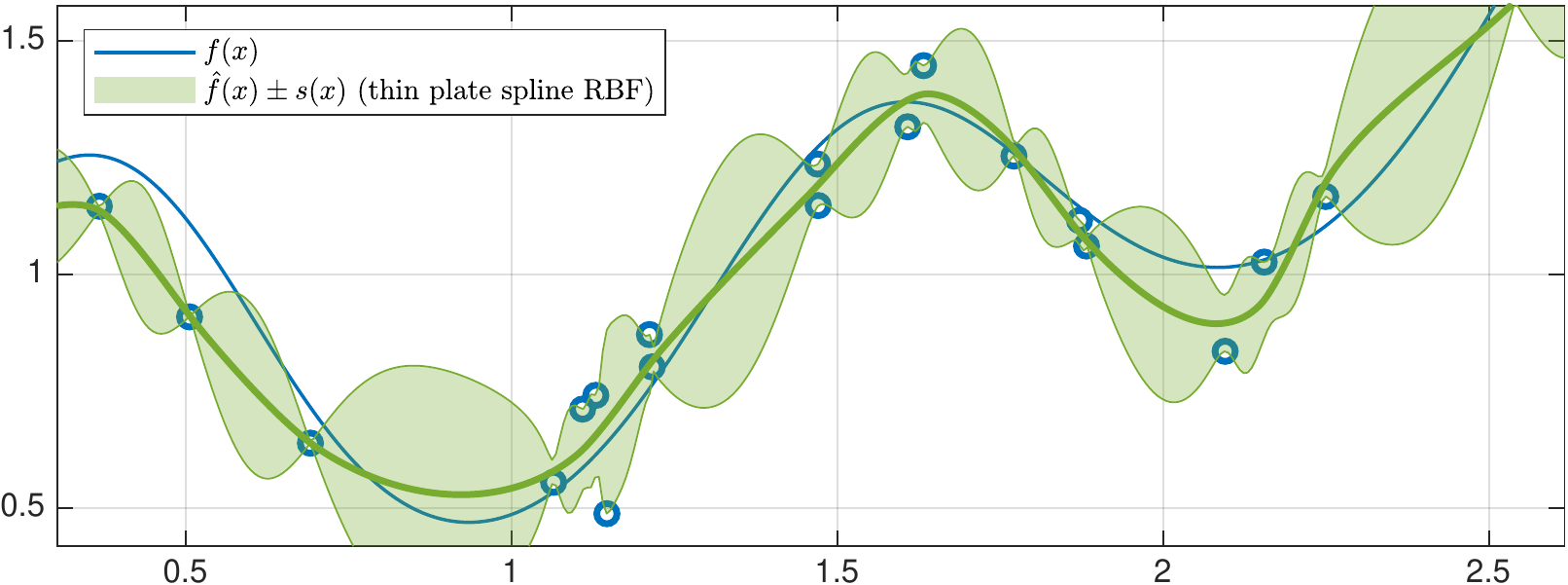}
\end{center}
\caption{Zoomed plot of $\hat f(x)\pm s(x)$ for the scalar example
as in Figure~\ref{fig:f_s_d} when 50 samples of $f(x)$ are measured
with noise $\varepsilon\sim{\mathcal N}(0,10^{-2})$ and $\epsilon_{\rm SVD}=10^{-2}$}
\label{fig:f_s_d_noise}
\end{figure}

In Bayesian optimization, such an exploration term is provided by the covariance associated with the Gaussian process. A function measuring ``bumpiness'' of a surrogate RBF function was used in~\cite{Gut01}. Here instead we propose two functions that provide
exploration capabilities, that can be used in alternative to each other or in a combined way. 
First, as suggested in~\cite{JK11} for IDW functions, 
we consider the confidence interval function $s:\rr^n\to\rr$ for $\hat f$ defined by
\begin{equation}
    s(x)=\sqrt{\sum_{i=1}^Nv_i(x)(f_i-\hat f(x))^2}
\label{eq:confidence}
\end{equation}
We will refer to function $s$ as the \emph{IDW variance function} associated 
with $(X,F)$. Clearly, when $\hat f(x_i)=f(x_i)$ then $s(x_i)=0$ for all $i=1,\ldots,N$ (no uncertainty at points $x_i$ where $f$ is evaluated exactly). See Figure~\ref{fig:f_s_d} for a noise-free example and Figure~\ref{fig:f_s_d_noise} for the case of noisy measurements of $f$.

\begin{figure}[t]
\begin{center}
\includegraphics[width=.8\hsize]{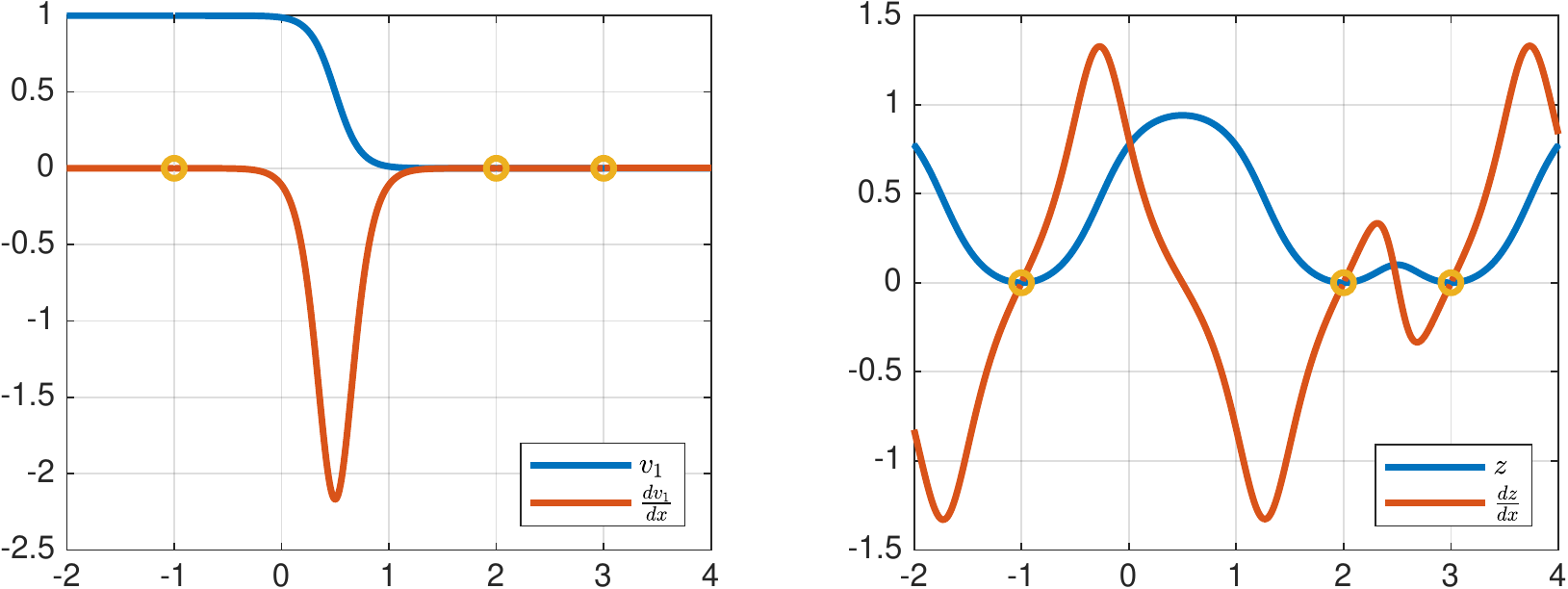}
\end{center}
\caption{A scalar example of function $v_1(x)$ and $z(x)$
for $x_j=j$, $j=1,2,3$}
\label{fig:v1_z}
\end{figure}

Second, we introduce the new \emph{IDW distance function}
$z:\rr^n\to\rr$ defined by
\begin{equation}
    z(x)=\left\{\ba{ll}
0 & \mbox{if}\ x\in\{x_1,\ldots,x_N\}\\
\frac{2}{\pi}\tan^{-1}\left(\frac{1}{\sum_{i=1}^Nw_i(x)}\right)&
\mbox{otherwise}\ea\right.
\label{eq:IDW-function}
\end{equation}
where $w_i(x)$ is given by either~\eqref{eq:w-IDW-basic} or~\eqref{eq:w-IDW}.
The rationale behind~\eqref{eq:IDW-function} is that $z(x)$ is zero
at sampled points and grows in between. The arc tangent
function in~\eqref{eq:IDW-function} avoids that $z(x)$ grows excessively
when $x$ is located far away from all sampled points.
Figure~\ref{fig:v1_z} shows a scalar
example of function $v_1$ and $z$. 

Given parameters $\alpha,\delta\geq 0$ and $N$ samples $(X,F)$,
we define the following \emph{acquisition function} $a:\rr^n\to\rr$
\begin{equation}
    a(x)=\hat f(x)-\alpha s(x)-\delta\Delta F z(x)
\label{eq:acquisition}
\end{equation}
where $\Delta F=\max\{\max_i\{f_i\}-\min_i\{f_i\},\epsilon_{\rm \Delta F}\}$ is the range of the observed samples $F$ and the threshold $\epsilon_{\rm \Delta F}>0$ is introduced
to prevent the case in which $f$ is not a constant function but, by chance,
all sampled values $f_i$ are equal. Scaling $z$ by $\Delta F$ ease the selection of the hyperparameter $\delta$, as the amplitude of $\Delta F z$ is comparable to that of $\hat f$.

As we will detail next, given $N$ samples $(X,F)$ a global minimum of the acquisition function~\eqref{eq:acquisition} is used to define the $(N+1)$-th sample $x_{N+1}$ by solving the global optimization problem
\begin{equation}
x_{N+1}=\arg\min_{\ell\leq x\leq u,\ x\in\XX
} a(x)
\label{eq:xNp1}
\end{equation}
The rationale behind choosing~\eqref{eq:acquisition} for acquisition is the following. The term $\hat f$ directs the search
towards a new sample $x_{N+1}$ where the objective function $f$ is expected to be optimal, assuming that $f$ and its surrogate $\hat f$
have a similar shape, and therefore allows a direct \emph{exploitation} of the samples $F$ already collected. The other two terms account instead for the \emph{exploration} of the feasible set with the hope of finding better values of $f$, with $s$ promoting areas in which $\hat f$ is more uncertain and $z$ areas that have not been explored yet. Both $s$ and $z$ provide exploration capabilities, but with an important difference: function $z$ is totally independent on the samples $F$ already collected and promotes a more uniform exploration, $s$ instead depends on $F$ and the surrogate $\hat f$. The coefficients $\alpha$, $\delta$ determine the exploitation/exploration tradeoff one desires to adopt.

\begin{figure}[t]
\begin{center}
\includegraphics[width=.9\hsize]{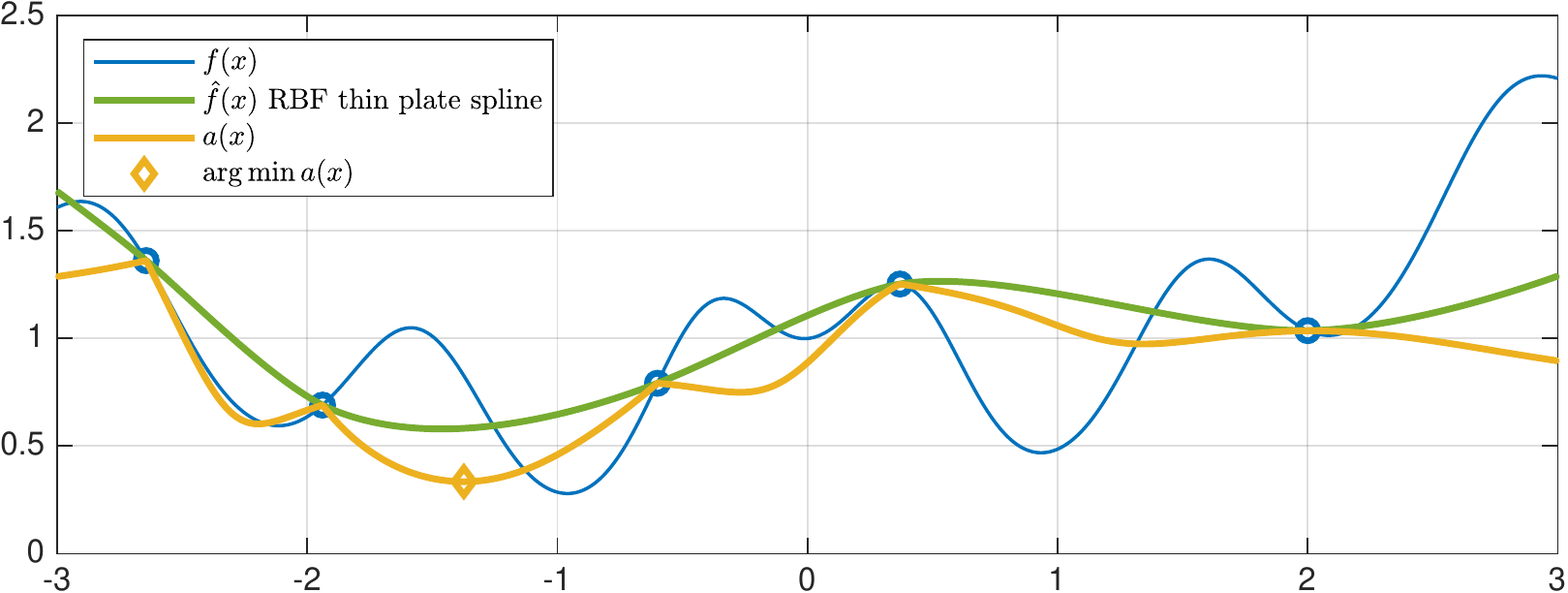}
\end{center}
\caption{Plot of $\hat f(x)$ and acquisition function $a(x)$ with $\alpha=1$, $\delta=\frac{1}{2}$, thin plate spline RBF with $\epsilon_{\rm SVD}=10^{-6}$,
for the scalar example
as in Figure~\ref{fig:rbf_vs_idw}, with $w_i(x)$ as in~\eqref{eq:w-IDW-basic} and $f$ as  in~\eqref{eq:f-1d-example}}
\label{fig:f_acquisition}
\end{figure}

For the example of scalar function $f$ in~\eqref{eq:f-1d-example} sampled at five random points, the acquisition function $a$ obtained by setting $\alpha=1$, $\delta=\frac{1}{2}$,
using a thin plate spline RBF with $\epsilon_{\rm SVD}=10^{-6}$,
and $w_i(x)$ as in~\eqref{eq:w-IDW-basic}, and 
the corresponding minimum are depicted in Figure~\ref{fig:f_acquisition}.

The following result, whose easy proof is reported in Appendix~\ref{app:proofs}, 
highlights a nice property of the acquisition function $a$:
\begin{lemma}
Function $a$ is differentiable everywhere on $\rr^n$.
\label{lemma:sz}
\end{lemma}

Problem~\eqref{eq:xNp1} is a global optimization problem whose objective function and constraints are very easy to evaluate. It can be solved very efficiently using various global optimization techniques, either derivative-free~\cite{RS13} or, if $\XX=\{x: g(x)\leq 0\}$ and $g$ is also differentiable, derivative-based. In case some components of vector $x$ are restricted to be integer,~\eqref{eq:xNp1} can be solved by mixed-integer programming. 

\section{Global optimization algorithm}
\label{sec:algorithm}
Algorithm~\ref{algo:idwgopt}, that we will refer to as GLIS (GLobal minimum using Inverse distance weighting and Surrogate radial basis functions), summarizes the proposed approach to solve the global optimization problem~\eqref{eq:glob-opt} using surrogate functions (either IDW or RBF) and the IDW acquisition function~\eqref{eq:acquisition}.

\begin{algorithm}[t]
\caption{GLIS -- Global optimization algorithm based on IDW-RBF surrogates}
\label{algo:idwgopt}
~~\textbf{Input}: Upper and lower bounds $(\ell,u)$, constraint set $\XX$;
number $N_{\rm init}$ of initial samples, number $N_{\rm max}\geq N_{\rm init}$ of maximum number of function 
evaluations; $\alpha,\delta\geq 0$, $\epsilon_{\rm SVD}>0$, $\epsilon_{\rm \Delta F}>0$.
\vspace*{.1cm}\hrule\vspace*{.1cm}
\begin{enumerate}[label*=\arabic*., ref=\theenumi{}]
\item Tighten $(\ell,u)$ as in~\eqref{eq:BBox-tighten};
\item Scale problem as in~\eqref{eq:scaling};
\item Set $N\leftarrow N_{\rm init}$;
\item \label{algo:latin} Generate $N$ random initial samples $X=[x_1\ \ldots\ x_N]'$ using Latin hypercube sampling~\cite{MBC79};
\item Compute $\hat f$ as in~\eqref{eq:rbf},~\eqref{eq:rbf_solve} (RBF function) or as in~\eqref{eq:fhat-idw} (IDW function);
\item \textbf{While} $N<N_{\rm max}$ \textbf{do}
\begin{enumerate}[label=\theenumi{}.\arabic*., ref=\theenumi{}.\arabic*]
\item \label{algo:acquisition} Compute acquisition function $a$ as in~\eqref{eq:acquisition};
\item \label{algo:globopt} Solve global optimization problem~\eqref{eq:xNp1} and get $x_{N+1}$;
\item $N\leftarrow N+1$;
\end{enumerate}
\item \textbf{End}.
\end{enumerate}
\vspace*{.1cm}\hrule\vspace*{.1cm}
~~\textbf{Output}: Global optimizer $x^*=x_{N_{\rm max}}$.
\end{algorithm}

As common in global optimization based on surrogate functions, in Step~\ref{algo:latin} 
\emph{Latin Hypercube Sampling} (LHS)~\cite{MBC79} is used to generate the initial set $X$
of samples in the given range $\ell,u$. Note that the generated initial points may not satisfy the inequality constraints 
$x\in\XX$. 
We distinguish between two cases:
\begin{enumerate}
    \item [$i$)] the objective function $f$ can be evaluated outside the feasible set $\FF$;
    \item [$ii$)] $f$ cannot be evaluated outside $\FF$.
\end{enumerate}
In the first case, initial samples of $f$ falling outside $\FF$ are still useful to 
define the surrogate function and can be therefore kept. In the second case, since $f$ cannot be 
evaluated at initial samples outside $\FF$, a possible approach is to generate more than $N_{\rm init}$ samples and discard the infeasible ones before evaluating $f$. For example,
the author of~\cite{Blo14} suggests the simple method reported in Algorithm~\ref{algo:cLHS}. This requires the feasible
set $\FF$ to be full-dimensional. In case of linear inequality constraints $\XX=\{x:\ Ax\leq b\}$, 
full-dimensionality of the feasible set $\FF$ can be easily checked by
computing the Chebychev radius $r_{\FF}$ of $\FF$ via the LP~\cite{BV04}
\begin{equation}
    \ba{rl}
    r_{\FF} = \max_{r,x} &r \\
    \st& A_ix\leq b_i-\|A_i\|_2r,\ i=1,\ldots,q\\
       & \ell_i+r\leq x_i\leq u_i-r,\ i=1,\ldots,n
    \ea 
\label{eq:ChebyRadius}
\end{equation}
where in~\eqref{eq:ChebyRadius} the subscript $i$ denotes the $i$th row (component)
of a matrix (vector). The polyhedron $\FF$ is full dimensional if and only if $r_{\FF}>0$.
Clearly, the smaller the ratio between the volume of $\FF$ 
and the volume of the bounding box $B_{\ell_g,u_g}$, the larger on average will be the
number of samples generated by Algorithm~\ref{algo:cLHS}.

Note that, in alternative to LHS, the IDW function~\eqref{eq:IDW-function} could be also used 
to generate $N_{\rm init}$ feasible points by solving
\[
    x_{N+1}=\max_{x\in\FF}z(x)
\]
for $N=1,\ldots,N_{\rm init}-1$, for any $x_1\in\FF$.

\begin{algorithm}[t]
\caption{Latin hypercube sampling with constraints}
\label{algo:cLHS}
~~\textbf{Input}: Upper and lower bounds $(\ell,u)$ for $x$ and inequality constraint function $g:\rr^n\to\rr^q$, defining a full dimensional set $\FF=\{x\in\rr^n:\ \ell\leq x\leq u, g(x)\leq 0\}$; number $N_{\rm init}$ of initial samples.
\vspace*{.1cm}\hrule\vspace*{.1cm}
\begin{enumerate}[label*=\arabic*., ref=\theenumi{}]
\item $N\leftarrow N_{\rm init}$; $N_k\leftarrow 0$;
\item \textbf{While}  $N_k<N_{\rm init}$ \textbf{do}
\begin{enumerate}[label=\theenumi{}.\arabic*., ref=\theenumi{}.\arabic*]
\item Generate $N$ samples using Latin hypercube sampling;
\item $N_k\leftarrow$ number of samples satisfying $x\in\XX$; 
\item If $N_k<N_{\rm init}$ then increase $N$ by setting
\[
    N\leftarrow\left\{\ba{ll}
    \lceil \min\{20,1.1\frac{N_{\rm init}}{N_k}\}N\rceil & \mbox{if}\ N_k>0\\
    20N & \mbox{otherwise}
    \ea\right.
\]
\end{enumerate}
\item \label{step:end} \textbf{End}.
\end{enumerate}
\vspace*{.1cm}\hrule\vspace*{.1cm}
~~\textbf{Output}: $N_{\rm init}$ initial samples $X=[x_1\ \ldots\ x_{N_{\rm init}}]'$
satisfying $\ell\leq x_i\leq u$, $x\in\XX$. 
\end{algorithm}

The examples reported in this paper use the Particle Swarm Optimization (PSO) algorithm~\cite{VV07} 
to solve problem~\eqref{eq:xNp1} at Step~\ref{algo:globopt}, although several other global optimization methods such as DIRECT~\cite{Jon09} or others~\cite{HN99,RS13}
could be used in alternative. Inequality constraints $\XX=\{x:\ g(x)\leq 0\}$
can be handled as penalty functions, for example by replacing~\eqref{eq:xNp1}
with 
\begin{equation}
    x_{N+1}=\arg\min_{\ell\leq x\leq u} a(x)+\rho\Delta F\sum_{i=1}^q\max\{g_i(x),0\}^2
\label{eq:xNp1-penalty}
\end{equation}
where in~\eqref{eq:xNp1-penalty} $\rho\gg 1$. 
Note that due to the heuristic involved in constructing function $a$, it is not crucial to find global solutions of very high accuracy when solving problem~\eqref{eq:xNp1}.

The exploration parameter $\alpha$ promotes visiting points in $[\ell,u]$ where the function surrogate has largest variance, $\delta$ promotes instead pure exploration independently on the surrogate function approximation, as it is only based on the sampled points $x_1,\ldots,x_N$ and their mutual distance. For example, if $\alpha=0$ and $\delta\gg1$ Algorithm~\ref{algo:idwgopt} will try to explore the entire feasible region, with consequent slower detection of points
$x$ with low cost $f(x)$. On the other hand, setting $\delta=0$ will make GLIS proceed only based on the function surrogate and its variance, that may lead to miss regions in $[\ell,u]$ where a global optimizer is located. For $\alpha=\delta=0$, GLIS will proceed based on pure minimization of $\hat f$ that, as observed earlier, can easily lead to converge away from a global optimizer.

\begin{figure}[t]
\begin{center}
\includegraphics[width=\hsize]{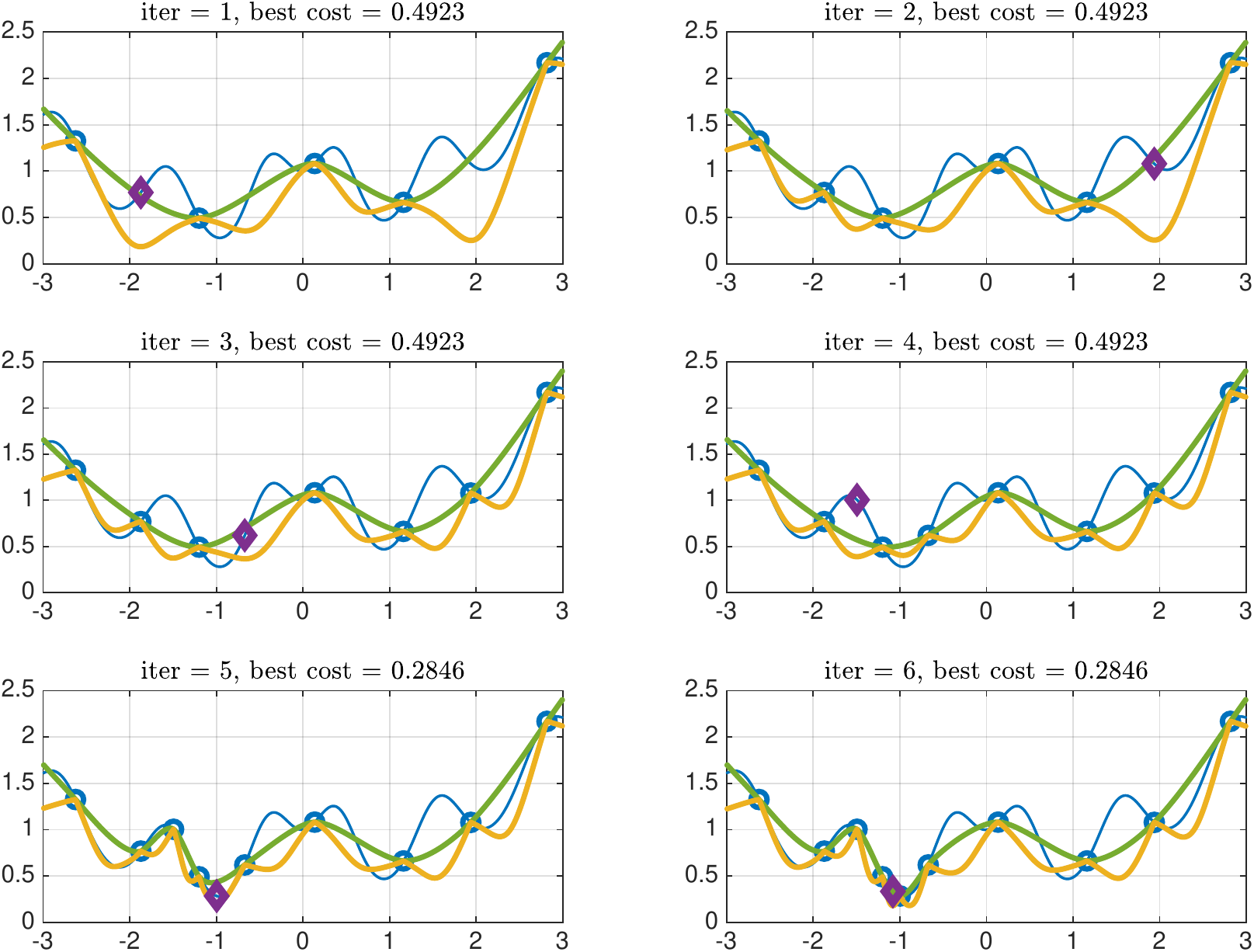}
\end{center}
\caption{GLIS steps 
    when applied to minimize the function $f$ given in~\eqref{eq:f-1d-example} using the same settings as in Figure~\ref{fig:f_acquisition} and $\epsilon_{\rm SVD}=10^{-6}$. The plots show function $f$ (blue), its samples $f_i$ (blue circles), the thin plate spline interpolation $\hat f$ with $\epsilon=0.01$ (green),
the acquisition function $a$ (yellow), and the minimum of the acquisition function reached at
$x_{N+1}$ (purple diamond)}
\label{fig:idwgopt-1d}
\end{figure}

Figure~\ref{fig:idwgopt-1d} shows the first six iterations of the GLIS
algorithm 
when applied to minimize the function $f$ given in~\eqref{eq:f-1d-example}
in $[-3,3]$ with $\alpha=1$, $\delta=0.5$.

\subsection{Computational complexity}
The complexity of Algorithm~\ref{algo:idwgopt}, as a function of the 
number $N_{\rm max}$ of iterations and dimension $n$ of the optimization space
and not counting the complexity of evaluating $f$, depends on Steps~\ref{algo:acquisition}
and~\ref{algo:globopt}. The latter depends on the global optimizer used to solve
Problem~\eqref{eq:xNp1}, which typically depends heavily on $n$. 
Step~\ref{algo:acquisition} involves 
computing $N_{\rm max}(N_{\rm max}-1)$ RBF values $\phi(\epsilon d(x_i,x_j))$,
$i,j=1,\ldots,N_{\rm max}$, $j\neq i$, compute the SVD decomposition of the $N\times N$
symmetric matrix $M$ in~\eqref{eq:RBF-system},
whose complexity is $O(N^3)$, and solve the linear
system in~\eqref{eq:RBF-system} ($O(N^2)$) 
at each step $N=N_{\rm init},\ldots,N_{\rm max}$.

\section{Numerical tests}
\label{sec:results}
In this section we report different numerical tests performed to assess the performance of the proposed algorithm (GLIS) and how it compares to Bayesian optimization (BO). For the latter,
we have used the off-the-shelf algorithm \texttt{bayesopt} implemented in the Statistics and Machine Learning Toolbox for MATLAB~\cite{SMLTBX}, based on the lower confidence bound as acquisition function. 
All tests were run on an Intel i7-8550 CPU @1.8GHz machine. Algorithm~\ref{algo:idwgopt} was run in MATLAB R2019b in interpreted code. The PSO solver~\cite{VV09} was used to solve problem~\eqref{eq:xNp1-penalty}.

\subsection{GLIS optimizing its own parameters}
\label{sec:self-optim}
We first use GLIS 
to optimize its own hyperparameters 
$\alpha$, $\delta$, $\epsilon$ when solving 
the minimization problem with $f(x)$ as in~\eqref{eq:f-1d-example} and $x\in[-3,3]$.
In what follows, we use the subscript $()_H$ to denote the parameters/function
used in the execution of the outer instance of Algorithm~\ref{algo:idwgopt} 
that is optimizing $\alpha$, $\delta$, $\epsilon$.
To this end, we solve the global optimization problem~\eqref{eq:glob-opt}
with $x=[\alpha\ \delta\ \epsilon]'$, $\ell_H=[0\ 0\ 0.1]'$, $u_H=[3\ 3\ 3]'$, and
\begin{equation}
f_H(x)=\sum_{i=1}^{N_t}\sum_{h=0}^{N_{\rm max}/2}(h+1)\min\{
f(x_{i,1}),\ldots,f(x_{i,N_{\rm max}/2+h})\}
\label{eq:f-selftuning}
\end{equation}
where $f$ is the scalar function in~\eqref{eq:f-1d-example}
that we want to minimize in $[-3,3]$, the $\min$ in~\eqref{eq:f-selftuning} provides the best objective value
found up to iteration $N_{\rm max}/2+h$, the term $(h+1)$ aims at penalizing
high values of the best objective the more the later they occur during the iterations,
$N_t=20$ 
is the number of times Algorithm~\ref{algo:idwgopt}
is executed to minimize $f_H$ for the same triplet $(\alpha,\delta,\epsilon)$,
$N_{\rm max}=20$ 
is the number of times $f$ is evaluated per execution,
$x_{i,N}$ is the sample generated by Algorithm~\ref{algo:idwgopt} during the $i$th run
at step $N$, $i=1,\ldots,N_t$, $N=1,\ldots,N_{\rm max}$.
Clearly~\eqref{eq:f-selftuning} penalizes failure to convergence close 
to the global optimum $f^*$ in $N_{\rm max}$ iterations without caring of how the algorithm performs during the first $N_{\rm max}/2-1$ iterations. 

\begin{figure}[t]
    \centerline{\includegraphics[width=0.8\hsize]{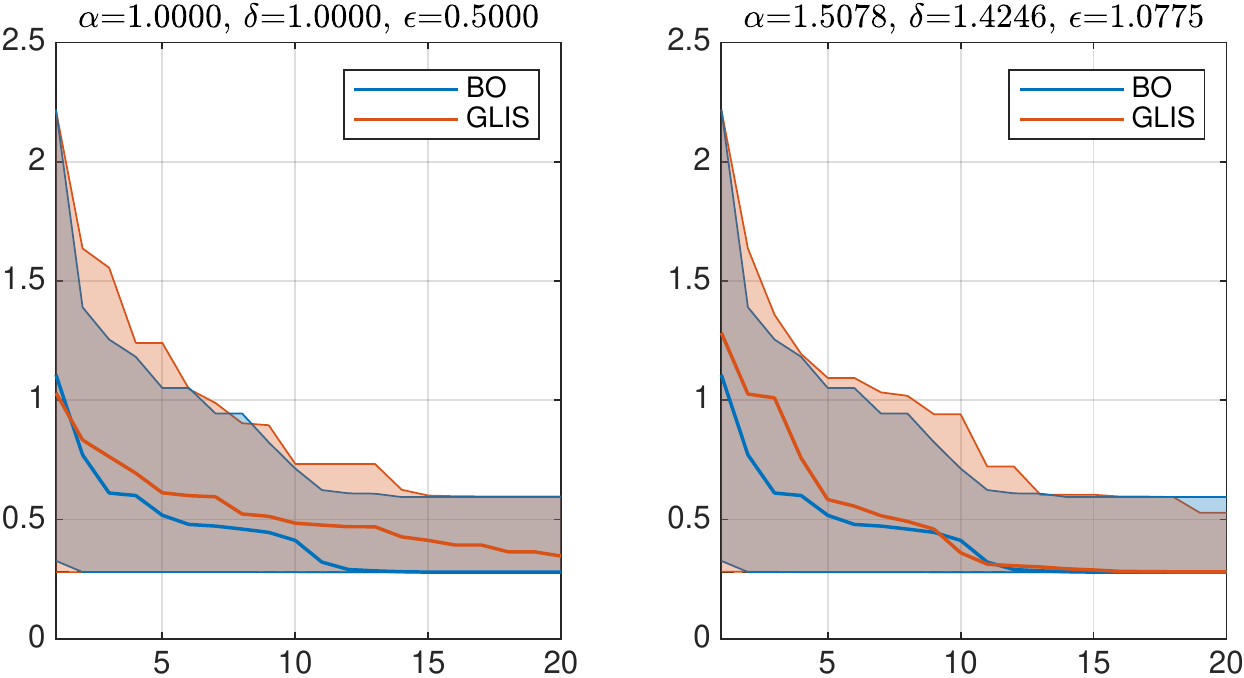}}
    \caption{Minimization of $f(x)$ as in~\eqref{eq:f-1d-example} in $[-3,3]$:
        tentative hyperparameters (left) and optimal hyperparameters (right)}
    \label{fig:optimized_params}
\end{figure}

In optimizing~\eqref{eq:f-selftuning}, the outer instance of Algorithm~\ref{algo:idwgopt} 
is run with $\alpha_H=1$, $\delta_H=0.5$, $\epsilon_H=0.5$, $N_{\rm init,H}=8$, $N_{\rm max,H}=100$,
and PSO as the global optimizer of the acquisition function. 
The RBF inverse quadratic function is used in both the inner and outer instances of Algorithm~\ref{algo:idwgopt}.
The resulting optimal selection is 
\begin{equation}
\alpha=0.8215,\ \delta=2.6788,\ \epsilon=1.3296
\label{eq:optimal-alpha-delta-epsil}
\end{equation}
Figure~\ref{fig:optimized_params} compares the behavior of GLIS (Algorithm~\ref{algo:idwgopt})
when minimizing $f(x)$ as in~\eqref{eq:f-1d-example} in $[-3,3]$ with tentative parameters $\alpha=1$, $\delta=1$, $\epsilon=0.5$ and with the optimal values in~\eqref{eq:optimal-alpha-delta-epsil}.

Clearly the results of the hyper-optimization depend on the function $f$ which is minimized in the inner loop. For a more comprehensive and general optimization of GLIS hyperparameters, one could alternatively consider in $f_H$ the average performance
with respect to a collection of problems instead of just one problem.

\subsection{Benchmark global optimization problems}
\label{sec:benchmarks}
We test the proposed global optimization algorithm on standard benchmark problems, summarized in Table~\ref{tab:benchmarks}. For each function the table shows the corresponding number of variables, upper and lower bounds, and the name of the example in~\cite{JY13} reporting the definition of the function. For lack of space, we will only consider the GLIS algorithm implemented
using inverse quadratic RBFs for the surrogate, leaving IDW only for exploration.

As a reference target for assessing the quality of the optimization results, for each benchmark problem the optimization algorithm DIRECT~\cite{Jon09} was used to compute the global optimum of the function
through the NLopt interface~\cite{NLopt}, except
for the \textsf{ackley} and \textsf{stepfunction2} benchmarks in which PSO is used instead due to the slow convergence of DIRECT on those problems. 

Algorithm~\ref{algo:idwgopt} is run by using the RBF inverse quadratic function with hyperparameters
obtained by dividing the values in~\eqref{eq:optimal-alpha-delta-epsil} by the number $n$ of variables,
with the rationale that exploration is more difficult in higher dimensions and it is therefore better to rely more on the surrogate function during acquisition.
The threshold $\epsilon_{\rm SVD}=10^{-6}$
is adopted to compute the RBF coefficients in~\eqref{eq:betaRBS}. The number of initial samples is $N_{\rm init}=2n$.

For each benchmark, the problem is solved $N_{\rm test}=100$ times to collect statistically  significant enough results. The last two columns of Table~\ref{tab:benchmarks} report
the average CPU time spent for solving the $N_{\rm test}=100$ instances 
of each benchmark using BO and GLIS. 
As the benchmark functions are very easy to compute, the CPU time spending
on evaluating the $N_{\rm max}$ function values $F$ is negligible, so the time
values reported in the Table are practically those due to the execution of the algorithms.
Algorithm~\ref{algo:idwgopt} (GLIS) is between 4.6 and 9.4 times faster than Bayesian optimization (about 7 times faster on average). The execution time of GLIS 
in Python 3.7 on the same machine, using the PSO package \texttt{pyswarm} (\url{https://pythonhosted.org/pyswarm})
to optimize the acquisition function, is similar to that of the BO package \texttt{GPyOpt}~\cite{gpyopt2016}.

\begin{table}
    {\small
        \begin{tabular}{l|c|c|c|l|r|r}
            benchmark & && &function name & \\
            problem &$n$ & $\ell$ & $u$& \cite{JY13} & BO[s] & GLIS [s]\\[.5em]\hline
            \textsf{ackley} & 2 & $\smallmat{-5\\-5}$ & $\smallmat{5\\5}$ &
            \emph{Ackley 1}, $D=2$ & 29.39 &  3.13
            \\[.5em]\hline
            \textsf{adjiman} & 2 & $\smallmat{-1\\-1}$ & $\smallmat{2\\1}$ & 
            \emph{Adjiman} &  3.29 &  0.68
            \\[.5em]\hline
            \textsf{branin} &2 & $\smallmat{-5\\0}$ & $\smallmat{10\\15}$ & 
            \emph{Branin RCOS} & 9.66 &  1.17
            \\[.5em]\hline
            \textsf{camelsixhumps} &2 & $\smallmat{-5\\-5}$ & $\smallmat{5\\5}$ & 
            \emph{Camel - Six Humps} &  4.82 &  0.62
            \\[.5em]\hline
            \textsf{hartman3} &3 & $[0\ 0\ 0]'$ & $[1\ 1\ 1]'$ & \emph{Hartman 3}
            & 26.27 &  3.35\\[.5em]\hline
            \textsf{hartman6} &6 & $x_i\geq 0$ & $x_i\leq 1$ &  
            \emph{Hartman 6} & 54.37 &  8.80\\[.5em]\hline
            \textsf{himmelblau} &2 & $\smallmat{-6\\-6}$ & $\smallmat{6\\6}$ & \emph{Himmelblau }  & 7.40 &  0.90 \\[.5em]\hline
            \textsf{rosenbrock8} &8 & $\smallmat{-30\\-30}$ & $\smallmat{30\\30}$ &  
            \emph{Rosenbrock 1}, $D=8$ & 63.09 & 13.73\\[.5em]\hline
            \textsf{stepfunction2} &4 & $x_i\geq -100$ & $x_i\leq 100$ & 
            \emph{Step 2}, $D=5$ &  11.72 &  1.81\\[.5em]\hline
            \textsf{styblinski-tang5} &5 &$x_i\geq -5$ & $x_i\leq 5$ & 
            \emph{Styblinski-Tang}, $n=5$ & 37.02 &  6.10\\\hline
        \end{tabular}
        \caption{Benchmark problems considered in the comparison.
            Last two columns: average CPU time spent on each benchmark
            for solving the $N_{\rm test}=100$ instances analyzed in Figure~\ref{fig:benchmarks}
            by Bayesian optimization (BO) and GLIS (Algorithm~\ref{algo:idwgopt})
            \label{tab:benchmarks}
        }
    }
\end{table}

\begin{figure}[p]
\includegraphics[width=\hsize]{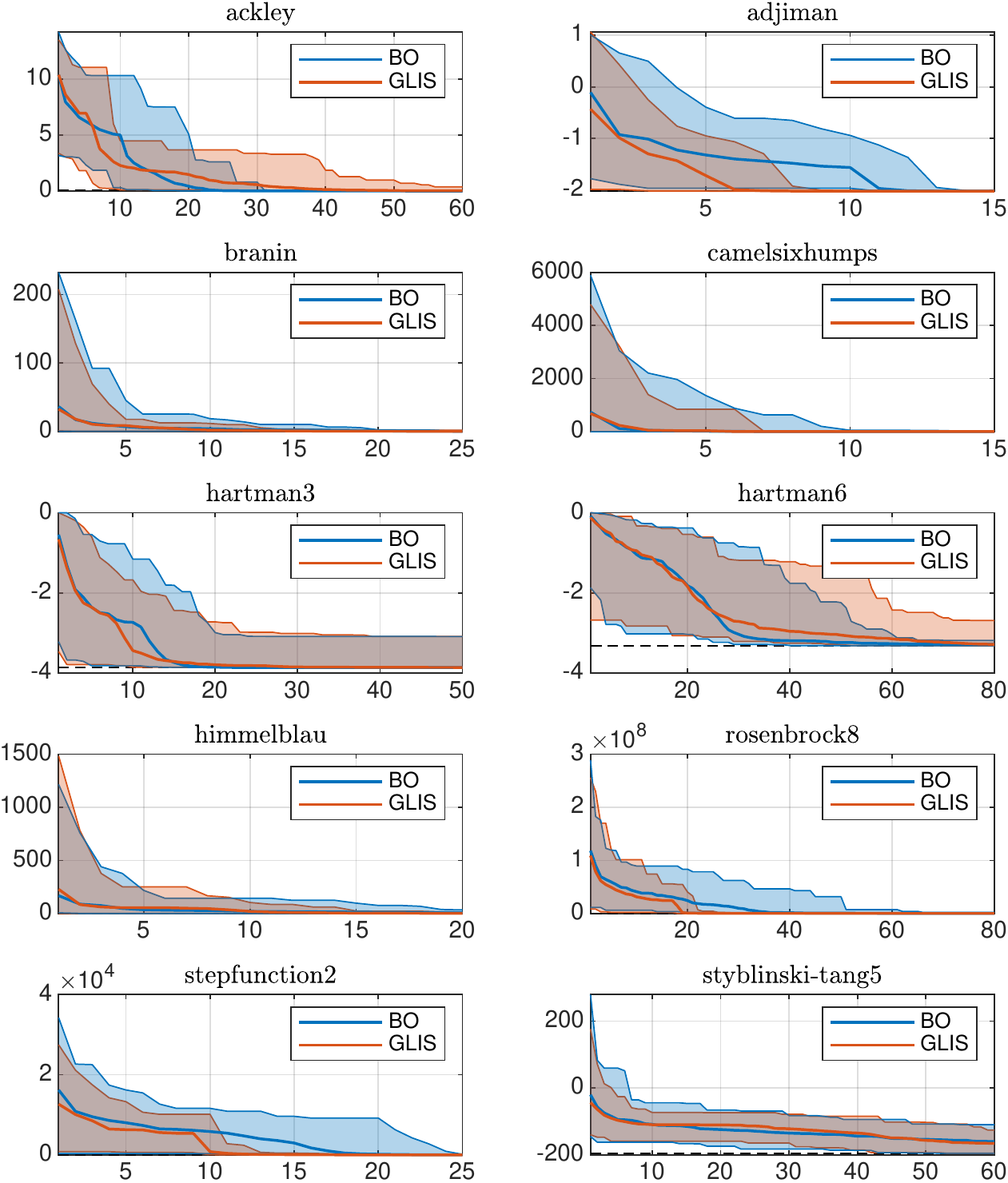}
\caption{Comparison between Algorithm~\ref{algo:idwgopt} (GLIS) and Bayesian optimization (BO) on benchmark problems. Each plot reports the average performance obtained over
$N_{\rm test}=100$ runs as a function of the number of function evaluations, along with the band defined by the best- and worst-case instances}
\label{fig:benchmarks}
\end{figure}

\begin{figure}[t]
\centerline{\includegraphics[width=\hsize]{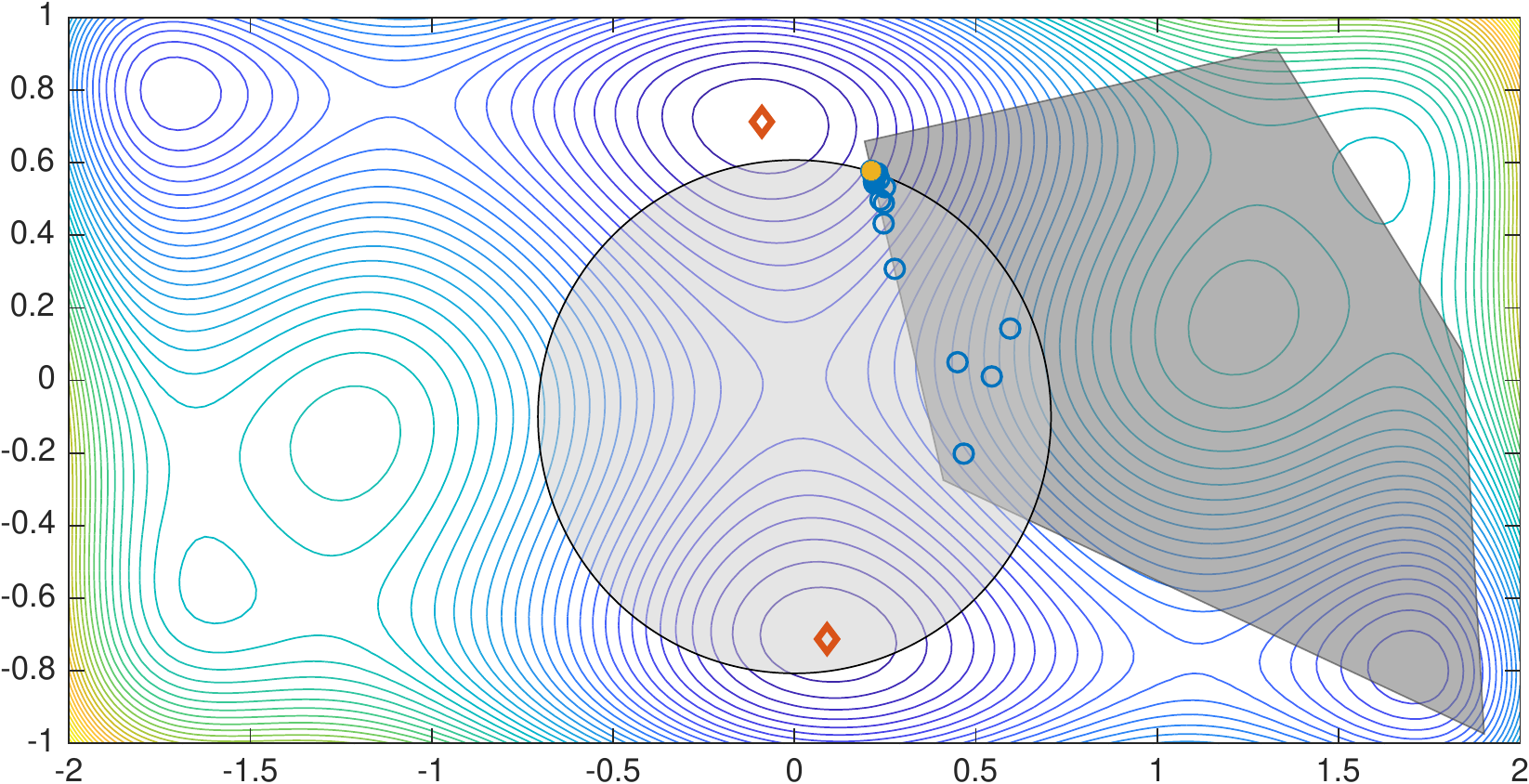}}
\caption{Constrained \textsf{camelsixhumps} problem: 
constrained minimum (yellow dot) and unconstrained global minima (red diamonds)}
\label{fig:camel_constr}
\end{figure}

In order to test the algorithm in the presence of constraints, we consider the
\textsf{camelsixhumps} problem and solve it under the following constraints
\[
     \ba{ll}
    -2\leq x_1\leq 2,\quad
-1\leq x_2\leq 1\\[1em]
            \smallmat{1.6295 & 1\\-1 & 4.4553\\-4.3023 & -1\\ -5.6905 &-12.1374\\17.6198 &1}x\leq \smallmat{3.0786\\ 2.7417\\ -1.4909\\ 1\\ 32.5198},&
    x_1^2+(x_2+0.1)^2\leq 0.5
    \ea
\]
Algorithm~\ref{algo:idwgopt} is run with hyperparameters set by dividing by $n=2$
the values obtained in~\eqref{eq:optimal-alpha-delta-epsil} 
and with $\epsilon_{\rm SVD}=10^{-6}$, $N_{\rm init}=2n$
for $N_{\rm max}=20$ iterations, with penalty $\rho=1000$ in~\eqref{eq:xNp1-penalty}.
The results are plotted in Figure~\ref{fig:camel_constr}.
The unconstrained two global minima of the function are located at $\smallmat{
-0.0898 \\ 0.7126}$, $\smallmat{0.0898\\-0.7126}$.

\subsection{ADMM hyperparameter tuning for QP}
\label{sec:ADMM}
The Alternating Direction Method of Multipliers (ADMM)~\cite{BPCPE11} is a popular
method for solving optimization problems such as the following convex 
Quadratic Program (QP)
\begin{equation}
    \ba{rl}
    \phi(\theta)=\min_z &\frac{1}{2}z'Qz+(c+F\theta)'z \\
    \st& Az\leq b+S\theta
    \ea
\label{eq:mpQP}
\end{equation}
where $z\in\rr^n$ is the optimization vector, $\theta\in\rr^p$ is a vector of parameters
affecting the problem, and $A\in\rr^{q\times n}$,  $b\in\rr^{q}$,  $S\in\rr^{q\times p}$,
and we assume $Q=Q'\succ 0$.
Problems of the form~\eqref{eq:mpQP} arise for example in model predictive control applications~\cite{Bem06,Bem15}, where $z$ represents a sequence of future control inputs
to optimize and $\theta$ collects signals that change continuously
at runtime depending on measurements and set-point values. ADMM can be used effectively to solve QP problems~\eqref{eq:mpQP}, see for example the
solver described in~\cite{BSMGBB17}. A very simple ADMM formulation for QP is summarized
in Algorithm~\ref{algo:qpadmm}.

\begin{figure}[t]
\centerline{\includegraphics[width=.9\hsize]{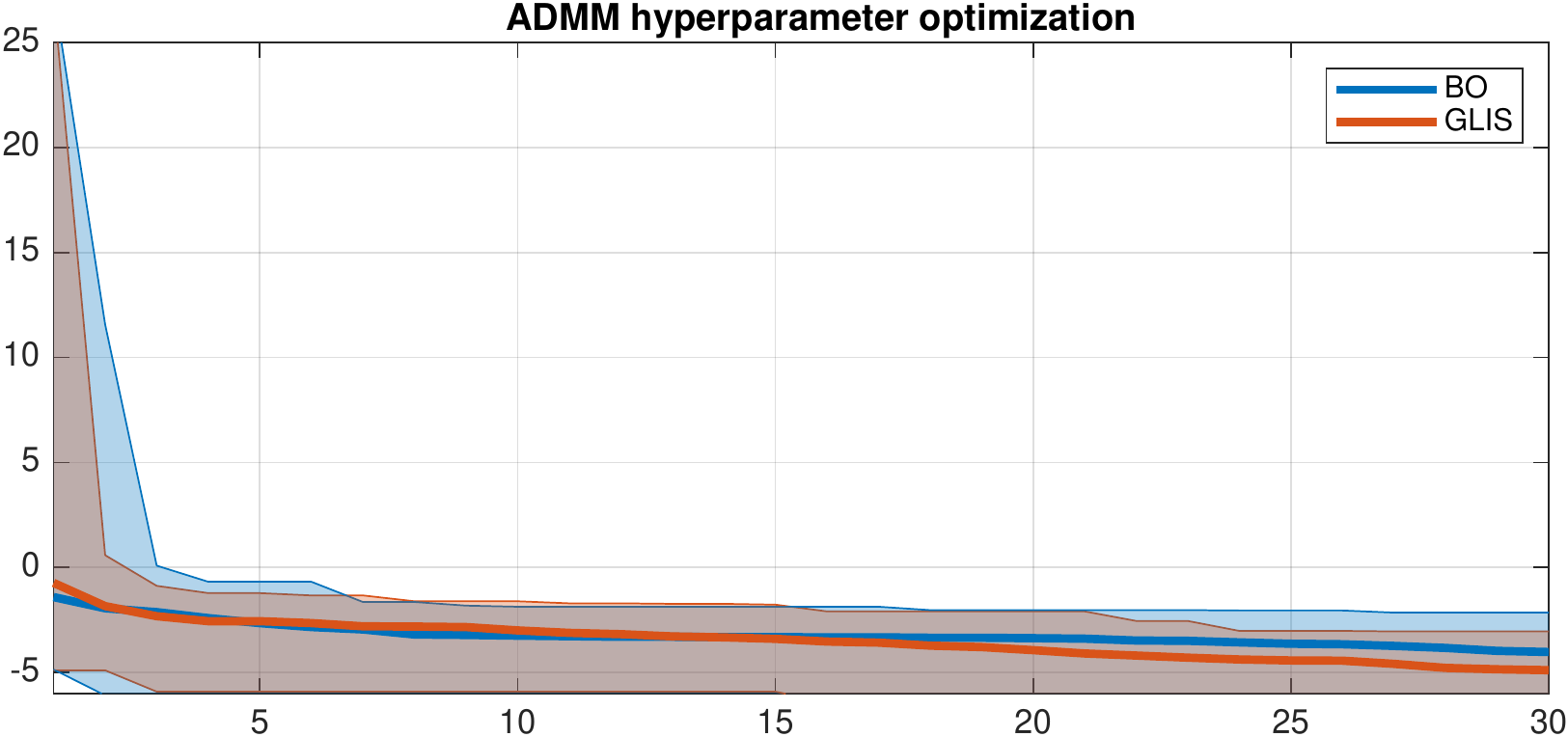}}
\caption{Hyperparameter optimization for ADMM}
\label{fig:results_ADMM}
\end{figure}

\begin{algorithm}[t]
\caption{ADMM for QP}
\label{algo:qpadmm}
~~\textbf{Input}: Matrices $Q,c,F,A,b,S$, parameter $\theta$, ADMM hyperparameters $\bar\rho,\bar\alpha$, number $N$ of ADMM iterations.
\vspace*{.1cm}\hrule\vspace*{.1cm}
\begin{enumerate}[label*=\arabic*., ref=\theenumi{}]
\item $M_A\leftarrow (\frac{1}{\bar\rho}Q+A'A)^{-1}A'$; $m_\theta\leftarrow (\frac{1}{\bar\rho}Q+A'A)^{-1}(c+F\theta)$; $b_\theta\leftarrow b+S\theta$;
\item $s\leftarrow 0$, $u\leftarrow 0$;
\item \textbf{for} $i=1,\ldots, N$ \textbf{do}:
\begin{enumerate}[label=\theenumi{}.\arabic*., ref=\theenumi{}.\arabic*]
\item $z\leftarrow M_A(s-u)-m_\theta$;
\item $w\leftarrow \bar\alpha Az+(1-\bar\alpha)s$;
\item $s\leftarrow \min(w+u,b_\theta)$;
\item $u\leftarrow u+w-s$;
\end{enumerate}
\item \textbf{End}.
\end{enumerate}
\vspace*{.1cm}\hrule\vspace*{.1cm}
~~\textbf{Output}: Optimal solution $z^*=z$.
\end{algorithm}
We consider a randomly generated QP test problem with $n=5$, $q=10$, $p=3$ that is feasible for all $\theta\in[-1,1]^3$, whose matrices are reported in Appendix~\ref{app:QPADMM} for reference.
We set $N=100$ in Algorithm~\ref{algo:qpadmm}, and generate $M=2000$
samples $\theta_i$ uniformly distributed in $[-1,1]^3$. The aim is to find the hyperparameters $x=[\bar\rho\ \bar\alpha]'$ that provide the best QP solution quality. This is expressed by the following objective function 
\beqar
    f(x) &=& \log\left(\frac{1}{M}\sum_{j=1}^M\max\left\{\frac{\phi^*_j(x)-\phi^*(x)}{1+|\phi^*(x)|},0\right\}\right.\nonumber\\&&\left.+
    \bar\beta\max\left\{\max_i\left\{\frac{A_iz_j^*(x)-b_i-S_ix}{1+|b_i+S_ix|}\right\},0\right\}\right)
\label{eq:ADMM-performance}
\eeqar
where $\phi^*_j(x)$, $z^*_j(x)$ are the optimal value and optimizer found at run $\#j$, respectively,
$\phi^*(x)$ is the solution of the QP problem obtained by running the very fast and accurate ODYS QP solver~\cite{ODYSQP}. The first term in~\eqref{eq:ADMM-performance} measures relative optimality, the second term relative violation of the constraints, and we set $\bar\beta=1$. Function $f$ in~\eqref{eq:ADMM-performance} is minimized for $\ell=\smallmat{0.01\\0.01}$ and $u=\smallmat{3\\3}$ using GLIS 
with the same parameters used in Section~\ref{sec:benchmarks} and, for comparison, by Bayesian optimization. The test is repeated $N_{\rm test}=100$ times and the results are depicted in Figure~\ref{fig:results_ADMM}. The resulting hyperparameter tuning 
that minimized the selected ADMM performance index~\eqref{eq:ADMM-performance} 
is $\bar\rho=0.1566$, $\bar\alpha=1.9498$. 

\section{Conclusions}
\label{sec:conclusions}
This paper has proposed an approach based on surrogate
functions to address global optimization problems whose objective
function is expensive to evaluate, possibly under 
constraints that are inexpensive to evaluate. Contrarily to Bayesian optimization methods, the approach is driven
by deterministic arguments based on radial basis functions to create
the surrogate, and on inverse distance weighting to characterize the uncertainty
between the surrogate and the black-box function to optimize, as well as
to promote the exploration of the feasible space. The computational burden
associated with the algorithm is lighter then the one of Bayesian optimization
while performance is comparable. 

Current research is devoted to extend the approach
to include constraints that are also expensive to evaluate, and to explore
if performance can be improved by adapting the parameters $\alpha$ and $\delta$ 
during the search.
Future research should address theoretical issues of convergence of the approach, by
investigating assumptions on the black-box function $f$ and on the parameters
$\alpha,\delta,\epsilon_{\rm SVD},\epsilon_{\rm \Delta F}$ of the algorithm, so
to allow guaranteeing convergence, for example using the arguments in~\cite{Gut01}
based on the results in~\cite{TZ89}.

\appendix

\section{Proofs}
\label{app:proofs}

\noindent\textbf{Proof of Lemma~\ref{lemma:fhat}}.
Property P1 easily follows from~\eqref{eq:v}. Property P2 also easily follows
from the fact that for all $x\in\rr^n$ the values $v_j(x)\in[0,1]$, $\forall j=1,\ldots,$ and $\sum_{j=1}v_j(x)=1$, so that 
\[
    \min_j\{f_j\}=\sum_{j=1}^Nv_j(x)\min_j\{f_j\}\leq\hat f(x)
\leq \sum_{j=1}^Nv_j(x)\max_j\{f_j\}=\max_j\{f_j\}
\]

Regarding differentiability, we first prove that for all $i=1,\ldots,N$, functions $v_i$ are differentiable everywhere on $\rr^n$, and that in particular $\nabla v_i(x_j)=0$ for all $j=1,\ldots,N$.
Clearly, functions $v_i$ are differentiable
for all $x\not\in\{x_1,\ldots,x_N\}$, $\forall i=1,\ldots,N$.
Let $e_h$ be the $h$th column of the identity matrix of order $n$.
Consider first the case in which $w_i(x)$ are given by~\eqref{eq:w-IDW}.
The partial derivatives of $v_i$ at $x_i$ are
\beqarno
    \frac{\partial v_i(x_i)}{\partial x_h}&=&\lim_{t\rightarrow 0}\frac{1}{t}
\left(\frac{w_i(x_i+te_h)}{\sum_{j=1}^Nw_j(x_i+te_h)}-1\right)
=\lim_{t\rightarrow 0}-\frac{\sum_{j\neq i}w_j(x_i+te_h)}{t\sum_{j=1}^Nw_j(x+te_h)}\\
&=&\lim_{t\rightarrow 0}-\frac{\sum_{j\neq i}w_j(x_i)}{t\frac{e^{-\|x_i+te_h-x_i\|^2}}{\|x_i+te_h-x_i\|^2}}
=\lim_{t\rightarrow 0}-t\frac{\sum_{j\neq i}w_j(x_i)}{e^{-t^2}}=0
\eeqarno
and similarly at $x_j$, $j\neq i$ are
\beqarno
    \frac{\partial v_i(x_j)}{\partial x_h}&=&\lim_{t\rightarrow 0}\frac{1}{t}
\left(\frac{w_i(x_j+te_h)}{\sum_{j=1}^Nw_j(x_j+te_h)}-0\right)=
\lim_{t\rightarrow 0}\frac{1}{t}\frac{w_i(x_j)}{\frac{e^{-t^2}}{t^2}+\sum_{k\neq j}w_k(x_j)}
=0
\eeqarno
In case $w_i(x)$ are given by~\eqref{eq:w-IDW-basic} differentiability follows
similarly, with $e^{-t^2}$ replaced by 1.
Therefore $\hat f$ is differentiable and
\[
    \nabla \hat f(x_j)=\sum_{i=1}^N f_i\nabla v_i(x_j)=0
\]
\hfill\QED

\noindent\textbf{Proof of Lemma~\ref{lemma:sz}}.
As by Lemma~\ref{lemma:fhat} functions $\hat f$ and $v_i$ are differentiable, $\forall i=1,\ldots,N$, it follows immediately that $s(x)$ is differentiable. Regarding differentiability of $z$, clearly it is differentiable
for all $x\not\in\{x_1,\ldots,x_N\}$, $\forall i=1,\ldots,N$.
Let $e_h$ be the $h$th column of the identity matrix of order $n$.
Consider first the case in which $w_i(x)$ are given by~\eqref{eq:w-IDW}.
The partial derivatives of $z$ at $x_i$ are
\beqarno
    \frac{\partial z(x_i)}{\partial x_h}&=&\lim_{t\rightarrow 0}\frac{1}{t}
\left(\frac{2}{\pi}\tan^{-1}\left(\frac{1}{\sum_{j=1}^Nw_j(x_i+te_h)}\right)-0\right)\\
&=&\lim_{t\rightarrow 0}
\frac{2}{\pi t}\tan^{-1}\left(\frac{1}{\frac{e^{-t^2}}{t^2}+\sum_{j\neq i}w_j(x_i)}\right)=\lim_{t\rightarrow 0}
\frac{2}{\pi t}\tan^{-1}\left(\frac{t^2}{1+\sum_{j\neq i}w_j(x_i)}\right) = 0
\eeqarno
In case $w_i(x)$ are given by~\eqref{eq:w-IDW-basic} differentiability follows
similarly, with $e^{-t^2}$ replaced by 1.
Therefore the acquisition function $a$ is differentiable for all $\alpha,\delta\geq 0$.
\hfill\QED

\section{Matrices of parametric QP considered in Section~\ref{sec:ADMM}}
\label{app:QPADMM}
\[
    \ba{l}
    Q=\footnotesize{\matrice{rrrrrrr}{ 
      &  6.6067  & -1.6361 &   2.8198 &   0.3776 &   3.1448\\
      &  -1.6361 &   0.9943  & -0.9998 &  -0.4786 &  -0.5198\\
      &  2.8198  & -0.9998  &  4.0749  &  0.2183  &  0.2714\\
      &  0.3776  & -0.4786  &  0.2183  &  0.7310  &  0.1689\\
      &  3.1448  & -0.5198 &   0.2714  &  0.1689  &  2.1716&}} \quad
    c=\footnotesize{\matrice{rrr}{  
      & -11.4795\\
      & 1.0487\\
      &  7.2225\\
      &  25.8549\\
      &  -6.6689&}} \\~\\
  A=\footnotesize{\matrice{rrrrrrr}{ 
          &-0.8637 &  -1.0891 &  -0.6156 &   1.4193 &  -1.0000\\
          &  0.0774  &  0.0326  &  0.7481  &  0.2916  & -1.0000\\
          &  -1.2141 &   0.5525 &  -0.1924 &   0.1978 &  -1.0000\\
          &  -1.1135 &   1.1006 &   0.8886 &   1.5877 &  -1.0000\\
          &  -0.0068 &   1.5442 &  -0.7648 &  -0.8045 &  -1.0000\\
          &  1.5326  &  0.0859  & -1.4023  &  0.6966  & -1.0000\\
          &  -0.7697 &  -1.4916 &  -1.4224 &   0.8351 & -1.0000\\
          &  0.3714  & -0.7423  &  0.4882  & -0.2437  & -1.0000\\
          &  -0.2256 &  -1.0616 &  -0.1774 &  0.2157  & -1.0000\\
          &  1.1174  &  2.3505  & -0.1961  & -1.1658  & -1.0000 &}} \quad
   b=\footnotesize{\matrice{rrr}{
           &  0.0838\\
           &  0.2290\\
           & 0.9133\\
           &  0.1524\\
           &  0.8258\\
           &  0.5383\\
           &  0.9961\\
           &  0.0782\\
           &  0.4427\\
           &  0.1067&}}\\~\\
    F=\footnotesize{\matrice{rrrrr}{
      &   1.8733 &   8.4038  & -6.0033\\
      &  -0.8249  & -8.8803  &  4.8997\\
      &  -19.3302 &   1.0009 &   7.3936\\
      &  -4.3897  & -5.4453  & 17.1189\\
      &  -17.9468 &   3.0352 &  -1.9412 &}}    
    \quad
     S=\footnotesize{\matrice{rrrrr}{&  2.9080 &  -0.3538  &  0.0229\\
       & 0.8252  & -0.8236 &  -0.2620\\
        &1.3790 &  -1.5771  & -1.7502\\
        &-1.0582  &  0.5080  & -0.2857\\
        &-0.4686 &   0.2820 &  -0.8314\\
       & -0.2725 &   0.0335 &  -0.9792\\
       & 1.0984 &  -1.3337  & -1.1564\\
       & -0.2779 &   1.1275 &  -0.5336\\
       & 0.7015  &  0.3502  & -2.0026\\
       & -2.0518 &  -0.2991 &   0.9642 &}}
    \ea
\]
\end{document}